\documentclass[sigconf,nonacm]{acmart}
\usepackage{multirow}

\AtBeginDocument{%
  }

\setcopyright{none}
\begin{document}

\title[Modality Head Decoupling and Regulation]{Decoupling and Regulating Modality Heads for Fake News Detection under Missing Modality}

\author{Kai Qian}
\affiliation{
  \department{School of Computer Science and Technology}
  \institution{Soochow University}
  \country{China}}
\authornote{These authors contributed equally to this work.}
\author{Dongjie Xu}
\affiliation{
  \institution{Soochow University}
  \country{China}}
\authornotemark[1]
\author{Weijie Shi}
\affiliation{
  \institution{Hong Kong University of Science and Technology}
  \country{China}}
\author{Yezhao Xu}
\affiliation{
  \department{Taotian Group}
  \institution{Alibaba Group}
  \country{China}}
\author{Jiaqi Wang}
\affiliation{
  \department{Computer Science and Engineering}
  \institution{The Chinese University of Hong Kong}
  \country{China}}
\author{Hanghui Guo}
\affiliation{
  \department{School of Computer Science and Engineering}
  \institution{Southeast University}
  \country{China}}
\author{Mengze Li}
\affiliation{
  \institution{Hangzhou Dianzi University}
  \country{China}}
\author{Hao Chen}
\affiliation{
  \department{FiT}
  \institution{Tencent}
  \country{China}}
\author{Ziyi Liu}
\affiliation{
  \institution{Soochow University}
  \country{China}}
\author{Jiajie Xu}
\affiliation{
  \department{School of Computer Science and Technology}
  \institution{Soochow University}
  \country{China}}
\renewcommand{\shortauthors}{Qian et al.}

\begin{abstract}
Multimodal news often loses its image or text as it spreads across Web platforms. Existing methods treat this as information loss, assuming that the original veracity label remains valid. This assumption fails for fake news reusing authentic material, whose remainder may be genuine. Detectors should therefore decouple modality cues and adapt their contributions to modality availability. However, coarse news labels provide little guidance for modality decoupling, while training on complete inputs can induce modality bias. To address these challenges, we propose Head-wise Modality Decoupling and Regulation (HMDR) for multimodal large language models. First, Modality Head Decoupling decouples modality representations through soft head grouping by attention preference. Second, Modality Head Regulation regulates head contributions through compensation for imbalanced modality attention. Finally, Modality Invariance Learning promotes invariance by aligning representations of each modality under replacement of the other. Experiments on two benchmarks show that HMDR improves detection under missing modalities while maintaining performance on complete inputs.
\end{abstract}



\maketitle

\section{Introduction}
The Web enables news to spread rapidly as it is reposted and shared across platforms\footnote{\textit{Relevance to the Web:} As news spreads across Web platforms, images or text are often lost through cropping, reposting, or platform restrictions. HMDR addresses this challenge by learning modality decoupled representations that remain effective when only one modality is present, enabling reliable verification of incomplete multimodal Web content. This aligns with the Web Mining, Multimedia and Multilingual Content Analysis track's focus on robust mining of multimedia Web content under real-world Web dissemination conditions.}. Fake news spreads in the same way and can reach a wide audience, with consequences ranging from public confusion during health crises \cite{naeem2020covid} to offline violence \cite{fisher2016pizzagate} and declining trust in the media \cite{ognyanova2020misinformation}. Since fake news often combines text with images, detectors built to counter it typically examine both modalities, using cross-modal fusion \cite{jin2017multimodal,singhal2019spotfake}, consistency and ambiguity modeling \cite{xue2021detecting,chen2022cross}, and, more recently, multimodal large language models (MLLMs) \cite{jin2024fake}. These detectors usually take for granted that both the text and the image are available.

In practice, however, multimodal news often loses its image or text as it spreads across platforms. This problem, commonly referred to as missing modality, has been widely studied in multimodal learning \cite{wu2024deep,zhan2025systematic}, where it is typically handled by reconstructing the absent modality \cite{wang2023distribution,zeng2024missing,ke2025knowledge} or by making fusion robust to incomplete inputs \cite{lee2023multimodal,xu2024leveraging}. Recent fake news detectors adopt similar strategies \cite{zeng2025imol,zhu2025adaptivevitbert}. All these methods treat a missing modality as a loss of information and assume that the original veracity label still holds for the remaining content.

\begin{figure}[t]
  \centering
  \includegraphics[width=\columnwidth]{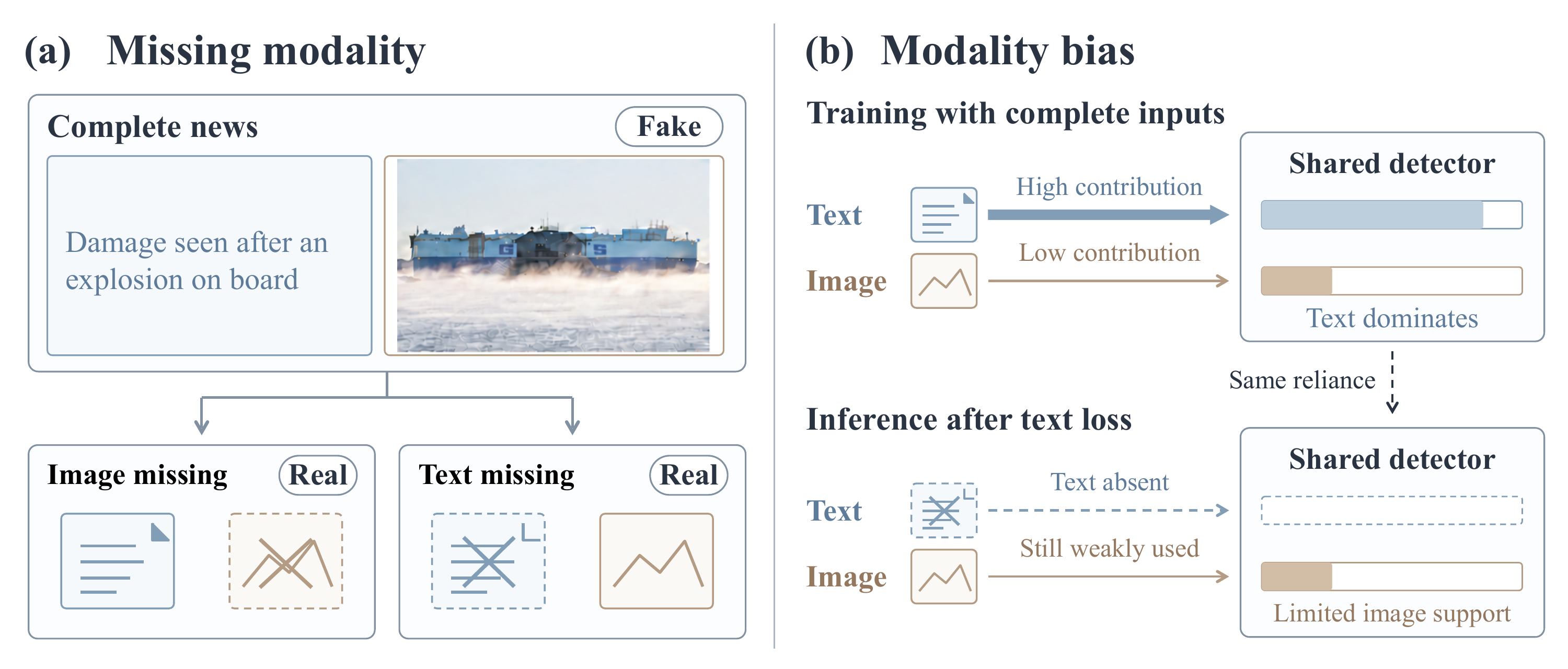}
  \caption{
  (a) The text and the image are each genuine, and the news is fake only because they are paired. Losing either one leaves real content.
  (b) A detector trained on complete inputs relies mainly on the text and keeps this reliance after the text is lost.
  }
  \label{fig:intro}
\end{figure}

This assumption breaks down when fake news is built from authentic material. Such news mixes authentic content with falsity that may lie in the text, in the image, or in how the two are combined \cite{shao2023detecting,guo2025each}. Once a modality is lost, the falsity may go with it, leaving genuine content behind. Figure~\ref{fig:intro}(a) shows such a case: the text and the image are each genuine, and the news is fake only because they are paired, so losing either one leaves real content. Rather than carrying over the original label, a detector should therefore decouple the evidence from each modality and adjust how much each contributes depending on which modalities are available. This is difficult for two reasons.

\textbf{First, coarse news labels provide little guidance for modality decoupling.}
A news label tells whether a news item as a whole is fake, but not where the falsity lies \cite{guo2025each}. Trained only on such labels, a detector receives no explicit signal for separating the evidence of each modality and may end up mixing them in shared representations. Modality-level labels would provide this signal, but they are costly to annotate and rarely available. Decoupling therefore has to be learned without them.

\textbf{Second, training on complete inputs can induce modality bias.}
When both modalities are always present during training, a detector may come to rely mostly on one of them, such as the text, and make little use of the other, as illustrated in Figure~\ref{fig:intro}(b). %
This bias carries over to inference. Once the text is lost, the image remains underused and offers little support for the final judgment. Modality contributions therefore need to be regulated according to which modalities are available.

To address these challenges, we draw on findings about attention heads. Prior studies show that attention heads specialize in distinct functions \cite{voita2019analyzing}, that some heads in the language models of MLLMs focus specifically on visual content \cite{bi2025unveiling}, and that adjusting individual heads can steer model behavior \cite{zhang2026locate}. Building on these findings, we propose Head-wise Modality Decoupling and Regulation (HMDR), which decouples modality representations and regulates their contributions at the level of attention heads in MLLMs. First, Modality Head Decoupling softly groups heads by whether they attend more to the text or to the image, so that modality representations can be decoupled without modality-level labels. Second, Modality Head Regulation adjusts head contributions to compensate for imbalanced modality attention, which mitigates the modality bias. Finally, Modality Invariance Learning aligns the representation of each modality before and after the other modality is replaced, so that it stays stable when the other modality is missing.

Our main contributions are as follows:
\begin{itemize}
    \item We point out that losing a modality can change the veracity of what remains of a Web news item, a problem that existing methods overlook by keeping the original news label.

    \item We propose HMDR, which decouples modality representations at the level of attention heads without modality-level labels and regulates head contributions to mitigate modality bias.

    \item Experiments on two benchmarks show that HMDR improves detection under missing modality while maintaining performance on complete inputs.
\end{itemize}

\section{Related Work}
\subsection{Multimodal Fake News Detection}
Multimodal fake news detection verifies news by jointly examining its text and images. Early methods extract features from each modality and fuse them for classification, for example through attention-based recurrent fusion \cite{jin2017multimodal} or simple concatenation \cite{singhal2019spotfake,segura2022multimodal}. Later work models finer interactions between the two modalities, using cross-modal attention \cite{tuan2021multimodal,ying2021multi}, co-attention \cite{wu2021multimodal}, bilinear pooling \cite{guo2023two}, and higher-order interaction terms \cite{sun2025multimodal}. Another line of work examines whether the text and the image agree, through consistency modeling \cite{xue2021detecting,shang2025semantic} and cross-modal ambiguity learning \cite{chen2022cross}. Contrastive learning has also been used to align the two modalities \cite{wang2023cross} or to learn more discriminative representations \cite{hua2023multimodal}. Further studies bring in external knowledge \cite{fu2023multimodal} and entity-level visual reasoning \cite{li2025entity}, and recent detectors build on large vision-language models \cite{jin2024fake}. These methods generally assume that both modalities are available at inference.

Only a few studies consider missing modality in fake news and rumor detection. IMOL reconstructs missing modalities for fake news video detection \cite{zeng2025imol}, while other methods compensate for the missing modality through knowledge distillation \cite{xu2023leveraging}, dynamic gating \cite{zhu2025adaptivevitbert}, or soft prompts \cite{chen2025trisprompt}. These methods mainly aim to recover or compensate for the lost information, implicitly assuming that the incomplete news shares the veracity of the complete one. As discussed in Section~1, this assumption overlooks that losing a modality can change the veracity of the remaining content.

\subsection{Multimodal Learning with Missing Modality}
Beyond fake news detection, missing modality has been widely studied in multimodal learning \cite{wu2024deep,zhan2025systematic}. One line of work recovers the missing modality from the available ones, by generating the absent data \cite{wang2024fedmmr,ke2025knowledge}, transferring its distribution \cite{wang2023distribution}, or predicting its features \cite{hoffman2016learning,zeng2024missing}. Another line avoids explicit recovery and instead makes the model robust to whichever modalities are present, through shared or low-rank representations \cite{liang2019learning,wang2020icmsc}, missing-aware attention and fusion \cite{ma2022multimodal,lee2023learning,ge2023metabev}, modality experts \cite{xu2024leveraging}, missing-aware prompts \cite{lee2023multimodal}, and probabilistic late fusion \cite{chen2022multimodal}.

Both lines aim to approach the performance obtained with complete inputs while keeping the original label. This goal is reasonable in tasks such as object detection \cite{chen2022multimodal}, emotion recognition \cite{xu2024leveraging}, and medical segmentation \cite{zeng2024missing}, where the label does not change when a modality is missing. In fake news detection, however, the missing modality may carry the falsity. Recovering it may bring the falsity back, and learning to predict the original label from the remainder may teach the detector to misjudge genuine content. HMDR therefore does not recover the missing modality, but decouples the evidence from each modality so that the judgment can follow the available content.

\subsection{Attention Heads in MLLMs}
Analyses of multi-head attention show that heads specialize in distinct functions, with a few specialized heads carrying most of the model's performance \cite{voita2019analyzing}. Similar specialization appears in MLLMs, where specific heads in the language model concentrate on visual content \cite{bi2025unveiling}. Beyond analysis, heads can be located and intervened on to steer model behavior \cite{li2023inference,zhang2026locate}. Existing head-level interventions in MLLMs \cite{he2025cracking,kang2025see} typically strengthen or redistribute the attention of image-related heads in frozen models at inference time, mainly to reduce hallucination. Our work differs in both purpose and setting. Rather than correcting a frozen model, HMDR uses the modality preference of heads to decouple modality representations and regulate their contributions during task fine-tuning, so that detection remains reliable under missing modality.

\section{Preliminaries}
\label{sec:preliminaries}

\begin{figure*}[t]
  \centering
  \includegraphics[width=\textwidth]{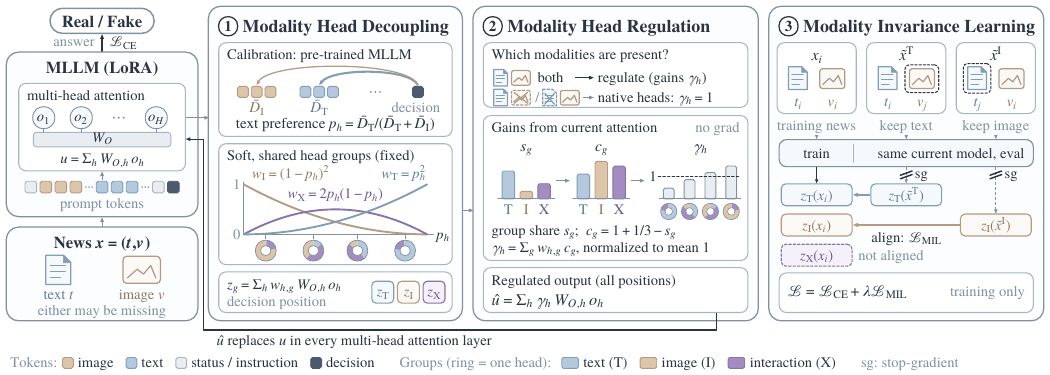}
  \caption{Overview of HMDR.
(1) MHD uses attention densities averaged over the calibration set to assign each head fixed, overlapping text, image, and interaction weights.
(2) MHR adjusts head contributions on complete inputs based on current attention, while missing-modality inputs retain native head outputs.
(3) During training, MIL aligns each modality representation with a stop-gradient reference from the same model, obtained by replacing the other modality.}
  \label{fig:framework}
\end{figure*}
\subsection{Task Definition}
Let $x=(t,v)$ denote a multimodal news item with text $t$ and image $v$. For training, we are given a set of complete news items $\mathcal{D}=\{(t_i,v_i,y_i)\}_{i=1}^{N}$, where the news label $y_i\in\{0,1\}$ indicates whether the news item as a whole is real ($0$) or fake ($1$). No modality-level labels are used for training.

At test time, a news item may arrive with only part of its content. We denote the available modalities by $\mathcal{M}$, which gives three views: the complete view $\mathcal{M}=\{\mathrm{T},\mathrm{I}\}$ (TI), the text-only view $\mathcal{M}=\{\mathrm{T}\}$ (T), and the image-only view $\mathcal{M}=\{\mathrm{I}\}$ (I). We write the input of a view as $x^{\mathcal{M}}$ and its label as $y^{\mathcal{M}}$, which indicates whether the available content is real or fake. For the complete view, $y^{\mathrm{TI}}=y$. For a missing view, $y^{\mathcal{M}}$ can differ from $y$, because the falsity may lie in the missing modality or in how the two modalities are combined. Our goal is to learn a detector from $\mathcal{D}$ that predicts $y^{\mathcal{M}}$ from $x^{\mathcal{M}}$ for all three views. Labels of the missing views are used only for evaluation.

\subsection{MLLM-based Detection}
We formulate detection as next-token prediction with an MLLM. The prompt first states whether the text and the image are present, then includes the available content, and ends with an instruction that asks the model to answer ``0'' for real or ``1'' for fake. We call the last position of the prompt the decision position and denote it by $n^{\ast}$. Let $s_0$ and $s_1$ be the logits of the two answer tokens at $n^{\ast}$. The probability that the input is fake is
\begin{equation}
P(\text{fake}\mid x)=\frac{\exp(s_1)}{\exp(s_0)+\exp(s_1)}.
\label{eq:prob}
\end{equation}
We use $\mathcal{T}$ and $\mathcal{V}$ to denote the positions of the news text tokens and the image tokens in the prompt. Either set is empty when the corresponding modality is missing.

\subsection{Attention Heads}
Each attention layer $l$ has $H$ heads. At position $n$, head $h$ attends to the positions $j\le n$ with weights $\alpha^{l}_{h,n,j}$ and produces an output $\mathbf{o}^{l}_{h,n}$.\footnote{Qwen3.5 applies an output gate to each head. We take $\mathbf{o}^{l}_{h,n}$ to be the gated output.} The output projection $W^{l}_{O}$ then combines all heads:
\begin{equation}
\mathbf{u}^{l}_{n}
=W^{l}_{O}\big[\mathbf{o}^{l}_{1,n};\dots;\mathbf{o}^{l}_{H,n}\big]
=\sum_{h=1}^{H}W^{l}_{O,h}\,\mathbf{o}^{l}_{h,n},
\label{eq:heads}
\end{equation}
where $W^{l}_{O,h}$ is the block of $W^{l}_{O}$ that corresponds to head $h$. Each head therefore adds its own term to the layer output, so its contribution can be weighted separately.

To describe how much a head attends to each modality, we use its attention at the decision position. The attention mass that head $h$ places on the text and on the image is
\begin{equation}
A^{l}_{h,\mathrm{T}}(x)=\sum_{j\in\mathcal{T}}\alpha^{l}_{h,n^{\ast},j},
\qquad
A^{l}_{h,\mathrm{I}}(x)=\sum_{j\in\mathcal{V}}\alpha^{l}_{h,n^{\ast},j}.
\label{eq:mass}
\end{equation}
Since the text and the image can differ greatly in the number of tokens, for a complete input we also use the attention density, which averages the mass over tokens: $D^{l}_{h,\mathrm{T}}(x)=A^{l}_{h,\mathrm{T}}(x)/|\mathcal{T}|$ and $D^{l}_{h,\mathrm{I}}(x)=A^{l}_{h,\mathrm{I}}(x)/|\mathcal{V}|$.

\section{Methodology}
\label{sec:method}

\subsection{Overview}
Figure~\ref{fig:framework} shows the framework of HMDR, which works on the attention heads of an MLLM detector fine-tuned with LoRA. Modality Head Decoupling (MHD) softly groups the heads into a text group, an image group, and an interaction group according to their attention preference, which yields a text representation and an image representation at each layer. Modality Head Regulation (MHR) adjusts head contributions on complete inputs to compensate for imbalanced modality attention. Modality Invariance Learning (MIL) trains the text and image representations to remain stable when the other modality is replaced. All three modules use only complete training news and news labels. We apply HMDR to every layer that uses standard multi-head attention and denote this set of layers by $\mathcal{S}$.\footnote{Qwen3.5-4B interleaves linear attention with standard attention. HMDR is applied to its 8 standard attention layers, and the linear attention layers are left unchanged.}

\subsection{Modality Head Decoupling}
Prior studies show that attention heads in MLLMs differ in how much they focus on visual content \cite{bi2025unveiling}. MHD uses this difference to group heads by modality. Before training, we run the pre-trained MLLM on a calibration set $\mathcal{C}$ of complete training news items, without their labels, and compute the text preference of each head:
\begin{equation}
p^{l}_{h}=\frac{\bar{D}^{l}_{h,\mathrm{T}}}{\bar{D}^{l}_{h,\mathrm{T}}+\bar{D}^{l}_{h,\mathrm{I}}},
\label{eq:pref}
\end{equation}
where $\bar{D}^{l}_{h,\mathrm{T}}$ and $\bar{D}^{l}_{h,\mathrm{I}}$ are the attention densities averaged over $\mathcal{C}$. A head with $p^{l}_{h}$ close to $1$ focuses on the text, a head close to $0$ focuses on the image, and a head near $0.5$ attends to both.

Instead of assigning each head to a single group, we give it a soft weight for each group:
\begin{equation}
w^{l}_{h,\mathrm{T}}=\big(p^{l}_{h}\big)^{2},\;\;
w^{l}_{h,\mathrm{I}}=\big(1-p^{l}_{h}\big)^{2},\;\;
w^{l}_{h,\mathrm{X}}=2p^{l}_{h}\big(1-p^{l}_{h}\big),
\label{eq:weights}
\end{equation}
where T, I, and X denote the text, image, and interaction groups. These weights are the three terms of $\big(p^{l}_{h}+(1-p^{l}_{h})\big)^{2}$ and thus sum to one for every head. A head with a clear preference contributes mainly to the group of that modality, and a head with a balanced preference contributes mainly to the interaction group. The preferences are computed once and then fixed, so MHD needs neither trainable parameters nor a threshold for assigning heads.

With these weights, the representation of group $g$ at layer $l$ is taken at the decision position:
\begin{equation}
\mathbf{z}^{l}_{g}(x)=\sum_{h=1}^{H}w^{l}_{h,g}\,W^{l}_{O,h}\,\mathbf{o}^{l}_{h,n^{\ast}},
\qquad g\in\{\mathrm{T},\mathrm{I},\mathrm{X}\}.
\label{eq:group}
\end{equation}
Because the weights of each head sum to one, the three groups together cover the contribution of every head. We use $\mathbf{z}^{l}_{\mathrm{T}}$ and $\mathbf{z}^{l}_{\mathrm{I}}$ as the text and image representations of layer $l$. They are computed from the head outputs before the regulation in Section~\ref{sec:mhr} and are the representations constrained by MIL (Section~\ref{sec:mil}). They do not enter the prediction path.

\subsection{Modality Head Regulation}
\label{sec:mhr}
When both modalities are present, the heads of one modality may take most of the attention at the decision position. The other modality may then contribute little to the prediction and receive a weaker training signal, which can lead to modality bias. MHR counteracts this by reweighting the heads on complete inputs.

For a complete input $x$, we first measure the attention that each group pays to its own modality at layer $l$:
\begin{equation}
\begin{gathered}
r^{l}_{\mathrm{T}}=\sum_{h=1}^{H}w^{l}_{h,\mathrm{T}}\,A^{l}_{h,\mathrm{T}},
\qquad
r^{l}_{\mathrm{I}}=\sum_{h=1}^{H}w^{l}_{h,\mathrm{I}}\,A^{l}_{h,\mathrm{I}},\\
r^{l}_{\mathrm{X}}=\sum_{h=1}^{H}w^{l}_{h,\mathrm{X}}\big(A^{l}_{h,\mathrm{T}}+A^{l}_{h,\mathrm{I}}\big),
\end{gathered}
\label{eq:share}
\end{equation}
where the interaction group is credited with its attention to both modalities. We normalize them into shares $s^{l}_{g}=r^{l}_{g}/\sum_{g'}r^{l}_{g'}$ and compute a compensation factor for each group:
\begin{equation}
c^{l}_{g}=1+\frac{1}{3}-s^{l}_{g}.
\label{eq:comp}
\end{equation}
A group that receives an equal share of $1/3$ keeps a factor of $1$. A group with a smaller share is amplified, and a group with a larger share is attenuated. Each head then combines the factors of the groups it belongs to:
\begin{equation}
\gamma^{l}_{h}=\frac{\tilde{\gamma}^{l}_{h}}{\frac{1}{H}\sum_{h'=1}^{H}\tilde{\gamma}^{l}_{h'}},
\qquad
\tilde{\gamma}^{l}_{h}=\sum_{g\in\{\mathrm{T},\mathrm{I},\mathrm{X}\}}w^{l}_{h,g}\,c^{l}_{g},
\label{eq:gain}
\end{equation}
where the normalization keeps the mean factor at one, so that only the relative contributions of the heads change. The regulated output replaces Eq.~\eqref{eq:heads} at every position:
\begin{equation}
\hat{\mathbf{u}}^{l}_{n}=\sum_{h=1}^{H}\gamma^{l}_{h}\,W^{l}_{O,h}\,\mathbf{o}^{l}_{h,n}.
\label{eq:regulated}
\end{equation}
The factors are computed for each input from the current attention and are treated as constants in backpropagation. Like MHD, MHR adds no trainable parameters.

MHR is applied whenever both modalities are available, in training and at inference. When one modality is missing, there is no other modality competing for attention, so we set $\gamma^{l}_{h}=1$ and keep the native head outputs. Since training uses complete news only, MHR is active throughout training, so that the less attended modality is not ignored as the detector learns.

\subsection{Modality Invariance Learning}
\label{sec:mil}
MHD groups heads by where they attend, but attending to the text does not guarantee that a representation depends only on the text. Hidden states mix information across positions, so text tokens placed after the image already carry image information from earlier layers. If the text representation still depends on the image, it shifts once the image is lost. MIL makes each modality representation stable to changes in the other modality, which encourages the two to be decoupled without modality-level labels.

For a training news item $x_i=(t_i,v_i)$, we take another training news item $x_j=(t_j,v_j)$ as its partner and build two replaced inputs: $\tilde{x}^{\mathrm{T}}_{i}=(t_i,v_j)$, which keeps the text and replaces the image, and $\tilde{x}^{\mathrm{I}}_{i}=(t_j,v_i)$, which keeps the image and replaces the text. If the text representation depends only on the text, it should be the same for $x_i$ and $\tilde{x}^{\mathrm{T}}_{i}$, and the same holds for the image representation with $\tilde{x}^{\mathrm{I}}_{i}$. We therefore minimize
\begin{equation}
\mathcal{L}_{\mathrm{MIL}}
=\frac{1}{2|\mathcal{B}||\mathcal{S}|}
\sum_{g\in\{\mathrm{T},\mathrm{I}\}}
\sum_{i\in\mathcal{B}}
\sum_{l\in\mathcal{S}}
\frac{\big\|\mathbf{z}^{l}_{g}(x_i)-\bar{\mathbf{z}}^{l}_{g,i}\big\|_2^{2}}
{\big\|\mathbf{z}^{l}_{g}(x_i)\big\|_2^{2}+\big\|\bar{\mathbf{z}}^{l}_{g,i}\big\|_2^{2}},
\label{eq:mil}
\end{equation}
where $\bar{\mathbf{z}}^{l}_{g,i}=\mathrm{sg}\big[\mathbf{z}^{l}_{g}(\tilde{x}^{g}_{i})\big]$ is the target, $\mathrm{sg}[\cdot]$ denotes stop-gradient, and $\mathcal{B}$ is the set of paired items in a mini-batch. We set $\mathcal{L}_{\mathrm{MIL}}=0$ when $\mathcal{B}$ is empty. Dividing by the energy of the two representations makes the loss insensitive to their scale, which varies across layers, and bounds each term between $0$ and $2$. For numerical stability, the denominator is clipped below at a small constant $\epsilon$. The targets $\bar{\mathbf{z}}^{l}_{g,i}$ are computed by the current model in evaluation mode without gradient, so no separate teacher is needed. The replaced inputs carry no labels and serve only as references. The interaction group is left unconstrained, since it is meant to capture how the two modalities relate.

We pair each training item with a fixed partner and keep the pair only if both replaced inputs have the same number of valid tokens as the original input. Items without such a partner contribute only to the classification loss.

\subsection{Training and Inference}
HMDR is trained with
\begin{equation}
\mathcal{L}=\mathcal{L}_{\mathrm{CE}}+\lambda\,\mathcal{L}_{\mathrm{MIL}},
\label{eq:total}
\end{equation}
where $\mathcal{L}_{\mathrm{CE}}$ is the cross-entropy of the answer token for the news label $y_i$ at the decision position, computed over the full vocabulary, and $\lambda$ balances the two terms. Only the LoRA parameters are updated. MHD and MHR add no parameters, and $\lambda$ is the only loss weight introduced by HMDR. Training uses complete news and news labels only.

At inference, the detector receives one of the three views. For the complete view, MHR regulates the heads as in training. For a missing view, the heads keep their native outputs. MIL and the replaced inputs are used only in training, so inference needs no extra forward pass. The detector predicts fake when $P(\text{fake}\mid x^{\mathcal{M}})$ in Eq.~\eqref{eq:prob} is at least $0.5$.

\section{Experiments}
\label{sec:exp}

We evaluate HMDR on two benchmarks to answer four research questions:
\textbf{RQ1}: Does HMDR improve detection when a modality is missing, while keeping its performance on complete news?
\textbf{RQ2}: How does each module of HMDR contribute?
\textbf{RQ3}: Does HMDR revise its judgment when the falsity of fake news is lost with the missing modality and only genuine content remains?
\textbf{RQ4}: Does MIL make the text and image representations less dependent on the other modality?

\subsection{Experimental Setup}
\label{sec:exp_setup}

\subsubsection{Datasets and view labels.}
We evaluate on FineFake \cite{zhou2026finefake} and the New York Times subset of MDSM \cite{zhang2026coherence}. For FineFake, we construct a 6:2:2 train/validation/test split at the item level, stratified by the original six categories with seed 42. For MDSM, we retain the official split and use a fixed subset of 20,000 training items; the validation and test splits are not subsampled.

View labels are derived from the original annotation categories. Genuine news is labeled real in both T and I. In FineFake, misleading combinations of genuine text and images also receive real labels in T and I. Text-fabrication and image-fabrication categories label only the fabricated modality as fake, leaving the other unknown; other fake-news categories leave both labels unknown. In MDSM, text-only manipulation gives fake text and a real image; facial-attribute manipulation gives real text and a fake image; face swapping gives a fake image but leaves the text label unknown. Joint manipulation gives fake labels in both T and I. An item is evaluated in a view only if the required input is present and the corresponding label is known.

Table~\ref{tab:data} reports the dataset statistics. In MDSM, one validation item and one test item have empty text but valid images with known labels, so they are evaluated only in I. Among fake news items in the test sets, FineFake contains 295 misleading combinations of genuine text and images, while MDSM contains 2,369 items with genuine text and a manipulated image and 1,671 with manipulated text and a genuine image. These cases test whether a detector can recognize genuine remaining content even when the complete news item is fake.

\begin{table}[t]
  \centering
  \caption{Dataset sizes. Test counts are real/fake for each eligible view. Train and Val list split sizes before view filtering; MDSM has 17,431 valid complete validation inputs.}
  \label{tab:data}
  \small
  \setlength{\tabcolsep}{2pt}
  \begin{tabular}{lrrccc}
    \toprule
    \multirow{2}{*}{Dataset} & \multirow{2}{*}{Train} & \multirow{2}{*}{Val} & \multicolumn{3}{c}{Test (real/fake)} \\
    \cmidrule(lr){4-6}
    & & & TI & T & I \\
    \midrule
    FineFake & 10,145 & 3,382 & 1,501/1,881 & 1,796/689 & 1,796/515 \\
    MDSM     & 20,000 & 17,432 & 8,824/8,609 & 11,193/3,552 & 10,495/6,939 \\
    \bottomrule
  \end{tabular}
\end{table}

\begin{table*}[t]
  \centering
  \caption{Test results (\%), using the label of each available view. TI, T, and I denote complete, text-only, and image-only inputs; F1 is macro-F1. Avg$_{\mathrm{uni}}$ averages T and I. Bold and underlining mark the best and second-best values among the listed methods. Zero-shot models are not fine-tuned; all trained variants use complete news and news labels only. MissRAG reuses its LoRA model on TI.}
  \label{tab:main}
  \small
  \setlength{\tabcolsep}{5pt}
  \begin{tabular}{l cc cc cc cc}
    \toprule
    \multirow{2}{*}{Method} & \multicolumn{2}{c}{TI} & \multicolumn{2}{c}{T} & \multicolumn{2}{c}{I} & \multicolumn{2}{c}{Avg$_{\mathrm{uni}}$} \\
    \cmidrule(lr){2-3}\cmidrule(lr){4-5}\cmidrule(lr){6-7}\cmidrule(lr){8-9}
    & Acc & F1 & Acc & F1 & Acc & F1 & Acc & F1 \\
    \midrule
    \multicolumn{9}{c}{\emph{FineFake}} \\
    \midrule
    MTS \cite{sun2025multimodal} & 80.04 & 79.92 & \underline{71.39} & 67.07 & 22.28 & 18.22 & 46.84 & 42.65 \\
    MIMoE-FND \cite{liu2025modality} & 80.46 & 80.03 & 67.32 & 65.94 & 22.46 & 18.47 & 44.89 & 42.21 \\
    DDU \cite{chen2026ddu} & 79.07 & 78.68 & 68.97 & 57.03 & 23.15 & 19.45 & 46.06 & 38.24 \\
    AOEPT \cite{lang2026aoept} & 77.62 & 76.98 & 60.28 & 60.04 & 22.93 & 19.15 & 41.61 & 39.60 \\
    DPL-NoPrompt \cite{lu2026dpl} & 76.76 & 75.44 & 65.47 & 64.01 & 38.99 & 38.90 & 52.23 & 51.45 \\
    \midrule
    Qwen3.5-4B Zero-shot & 54.73 & 37.20 & 39.84 & 38.88 & 23.28 & 19.85 & 31.56 & 29.36 \\
    Qwen3.5-4B LoRA & \underline{81.76} & 81.58 & 71.07 & 68.28 & 55.65 & 52.98 & 63.36 & 60.63 \\
    Qwen3.5-4B LoRA + MissRAG & \underline{81.76} & 81.58 & 68.65 & 66.30 & 63.26 & 60.57 & 65.96 & 63.44 \\
    Gemma4-E4B Zero-shot & 66.68 & 66.67 & 68.21 & 64.94 & \textbf{73.09} & 52.61 & \underline{70.65} & 58.78 \\
    Gemma4-E4B LoRA & \textbf{82.05} & \textbf{81.85} & 70.87 & \underline{69.02} & 68.50 & 60.71 & 69.68 & 64.86 \\
    Gemma4-E4B LoRA + MissRAG & \textbf{82.05} & \textbf{81.85} & 70.78 & 68.88 & 63.95 & \textbf{61.08} & 67.37 & \underline{64.98} \\
    \midrule
    HMDR (Ours) & \underline{81.76} & \underline{81.73} & \textbf{77.14} & \textbf{73.94} & \underline{72.61} & \underline{61.07} & \textbf{74.88} & \textbf{67.51} \\
    \midrule
    \multicolumn{9}{c}{\emph{MDSM}} \\
    \midrule
    MTS \cite{sun2025multimodal} & 79.54 & 79.53 & 94.86 & 93.22 & 74.91 & 73.30 & 84.88 & 83.26 \\
    MIMoE-FND \cite{liu2025modality} & 89.46 & 89.35 & \underline{95.45} & \underline{93.71} & 81.11 & 80.05 & \underline{88.28} & \underline{86.88} \\
    DDU \cite{chen2026ddu} & 86.47 & 86.35 & 84.14 & 70.92 & 55.89 & 55.22 & 70.02 & 63.07 \\
    AOEPT \cite{lang2026aoept} & 89.32 & 89.24 & 91.82 & 88.12 & \underline{81.38} & \underline{80.47} & 86.60 & 84.30 \\
    DPL-NoPrompt \cite{lu2026dpl} & 65.67 & 64.15 & 65.07 & 62.66 & 48.35 & 45.42 & 56.71 & 54.04 \\
    \midrule
    Qwen3.5-4B Zero-shot & 44.60 & 39.63 & 25.98 & 22.52 & 41.98 & 35.12 & 33.98 & 28.82 \\
    Qwen3.5-4B LoRA & 87.36 & 87.32 & 71.22 & 69.59 & 76.57 & 75.82 & 73.90 & 72.71 \\
    Qwen3.5-4B LoRA + MissRAG & 87.36 & 87.32 & 60.27 & 59.54 & 64.70 & 64.70 & 62.49 & 62.12 \\
    Gemma4-E4B Zero-shot & 49.62 & 48.01 & 46.84 & 44.43 & 59.25 & 41.48 & 53.05 & 42.95 \\
    Gemma4-E4B LoRA & \textbf{90.25} & \textbf{90.14} & 93.19 & 91.34 & 79.26 & 78.37 & 86.23 & 84.85 \\
    Gemma4-E4B LoRA + MissRAG & \textbf{90.25} & \textbf{90.14} & 57.30 & 55.97 & 69.79 & 69.65 & 63.54 & 62.81 \\
    \midrule
    HMDR (Ours) & \underline{90.01} & \underline{89.90} & \textbf{98.00} & \textbf{97.26} & \textbf{83.99} & \textbf{82.61} & \textbf{91.00} & \textbf{89.94} \\
    \bottomrule
  \end{tabular}
\end{table*}

\subsubsection{Evaluation protocol.}
All trainable variants learn from complete news and news labels, without view labels or modality removal during training. Zero-shot models receive no task-specific updates. Each trained variant selects its checkpoint by macro-F1 on the complete validation view and uses that checkpoint for TI, T, and I. Predictions are scored against the corresponding view label, with a fixed fake-probability threshold of $0.5$. We report accuracy (Acc), macro-F1 (F1), and their unweighted averages over T and I, denoted Avg$_{\mathrm{uni}}$. Macro-F1 is the primary metric because view classes are imbalanced: always predicting real on FineFake I yields 77.72\% accuracy but only 43.73\% macro-F1. Results are reported at seed 42.

\subsubsection{Baselines.}
We compare three groups of methods. \emph{Multimodal fake news detectors} include MTS \cite{sun2025multimodal}, MIMoE-FND \cite{liu2025modality}, and DDU \cite{chen2026ddu}, which use high-order interactions, modality-aware experts, and uncertainty-aware retrieval and fusion, respectively. \emph{Missing-modality methods} include AOEPT \cite{lang2026aoept} and DPL-NoPrompt \cite{lu2026dpl}. We adapt both to complete-input training and view-specific evaluation.

\emph{MLLM baselines} use Qwen3.5-4B \cite{qwen2026qwen35} and Gemma4-E4B \cite{gemmateam2026gemma4}, either zero-shot or fine-tuned with LoRA \cite{hu2022lora}. We also evaluate a text--image adaptation of MissRAG \cite{pipoli2025missrag} on each LoRA model: it retrieves the top training item using the available modality and supplies the retrieved missing modality as a prototype. Retrieval libraries contain selected training items only. MissRAG leaves complete inputs unchanged, so its TI scores equal those of the corresponding LoRA model. All methods share the same data splits and view-specific evaluation sets.

\subsubsection{Implementation.}
HMDR uses Qwen3.5-4B. The prompt indicates modality availability and asks for ``0'' (real) or ``1'' (fake), with thinking disabled. We truncate news text to 768 tokens and use an image-size budget of 65,536--262,144 pixels. The vision encoder is frozen; LoRA is applied to all language-model linear layers with rank 32, $\alpha=64$, and dropout 0.05, giving 64.93M trainable parameters. HMDR adds no trainable parameters. We train for five epochs using AdamW \cite{loshchilov2019decoupled}, learning rate $5\times10^{-5}$, weight decay 0.01, a cosine schedule with 10\% warm-up, batch size 6, and gradient clipping at 1.0. Backbone weights use bfloat16 and LoRA parameters use float32. The Qwen LoRA baseline shares these settings.

HMDR operates on Qwen3.5-4B's eight full-attention layers, each with 16 query heads. MHD estimates fixed head weights from 128 complete training items. We use $\lambda=0.01$ on both datasets. MIL requires replacement partners with compatible token positions, attention masks, and image grids, as well as equal sequence lengths. The matching procedure covers 5,852 of 10,145 FineFake items (57.7\%) and 13,778 of 20,000 MDSM items (68.9\%) per epoch; unmatched items contribute only to classification.

\subsection{Main Results (RQ1)}
\label{sec:exp_main}

\begin{table*}[t]
  \centering
  \caption{Ablation results (\%). F1 is macro-F1. Training variants use the same complete-view checkpoint-selection rule. Shuffling changes only the head-group weights used by MIL. ``MHR on missing views'' changes inference on the full model's fixed checkpoint, without retraining.}
  \label{tab:ablation}
  \small
  \setlength{\tabcolsep}{4.5pt}
  \begin{tabular}{l ccccc ccccc}
    \toprule
    \multirow{2}{*}{Variant} & \multicolumn{5}{c}{FineFake} & \multicolumn{5}{c}{MDSM} \\
    \cmidrule(lr){2-6}\cmidrule(lr){7-11}
    & TI F1 & T F1 & I F1 & Avg$_{\mathrm{uni}}$ F1 & Avg$_{\mathrm{uni}}$ Acc & TI F1 & T F1 & I F1 & Avg$_{\mathrm{uni}}$ F1 & Avg$_{\mathrm{uni}}$ Acc \\
    \midrule
    HMDR & 81.73 & 73.94 & 61.07 & \textbf{67.51} & 74.88 & 89.90 & 97.26 & 82.61 & \textbf{89.94} & \textbf{91.00} \\
    \midrule
    w/o MIL & 83.11 & 70.74 & 52.28 & 61.51 & 64.12 & 89.48 & 72.45 & 81.54 & 77.00 & 78.57 \\
    w/o MHR & 81.86 & 71.45 & 51.76 & 61.61 & 64.52 & 90.11 & 87.58 & 78.54 & 83.06 & 84.60 \\
    Shuffled groups in MIL & 82.01 & 73.42 & 58.98 & 66.20 & 75.43 & 88.43 & 19.47 & 76.83 & 48.15 & 50.88 \\
    MHR on missing views & 81.73 & 74.10 & 52.67 & 63.39 & \textbf{77.23} & 89.90 & 96.96 & 79.32 & 88.14 & 89.13 \\
    \bottomrule
  \end{tabular}
\end{table*}

\begin{table*}[t]
  \centering
  \caption{Accuracy (\%) by news label and remaining-view label. Each cohort contains items eligible in both TI and the indicated missing view; $n$ is its size. All models use Qwen3.5-4B. MDSM I excludes one item with empty text, giving 6,938 fake images here versus 6,939 in Table~\ref{tab:data}.}
  \label{tab:cohort}
  \small
  \setlength{\tabcolsep}{3.5pt}
  \begin{tabular}{l ccc ccc ccc ccc}
    \toprule
    & \multicolumn{3}{c}{FineFake T} & \multicolumn{3}{c}{FineFake I} & \multicolumn{3}{c}{MDSM T} & \multicolumn{3}{c}{MDSM I} \\
    \cmidrule(lr){2-4}\cmidrule(lr){5-7}\cmidrule(lr){8-10}\cmidrule(lr){11-13}
    News$\rightarrow$Remaining & LoRA & w/o MIL & HMDR & LoRA & w/o MIL & HMDR & LoRA & w/o MIL & HMDR & LoRA & w/o MIL & HMDR \\
    \midrule
    Real$\rightarrow$Real & \multicolumn{3}{c}{\emph{$n$=1,501}} & \multicolumn{3}{c}{\emph{$n$=1,501}} & \multicolumn{3}{c}{\emph{$n$=8,824}} & \multicolumn{3}{c}{\emph{$n$=8,824}} \\
    & 79.28 & 79.95 & \textbf{84.21} & 56.30 & 55.16 & \textbf{84.94} & 62.27 & 66.43 & \textbf{98.63} & 90.66 & 95.42 & \textbf{99.04} \\
    \midrule
    Fake$\rightarrow$Real & \multicolumn{3}{c}{\emph{$n$=295}} & \multicolumn{3}{c}{\emph{$n$=295}} & \multicolumn{3}{c}{\emph{$n$=2,369}} & \multicolumn{3}{c}{\emph{$n$=1,671}} \\
    & 20.68 & 23.73 & \textbf{44.07} & 24.75 & 29.15 & \textbf{65.42} & 61.84 & 68.30 & \textbf{99.20} & 12.81 & 56.85 & \textbf{61.94} \\
    \midrule
    Fake$\rightarrow$Fake & \multicolumn{3}{c}{\emph{$n$=689}} & \multicolumn{3}{c}{\emph{$n$=515}} & \multicolumn{3}{c}{\emph{$n$=3,552}} & \multicolumn{3}{c}{\emph{$n$=6,938}} \\
    & 74.75 & \textbf{79.68} & 75.91 & \textbf{71.46} & 69.51 & 40.78 & \textbf{99.72} & 98.51 & 95.64 & \textbf{74.01} & 72.69 & 70.16 \\
    \bottomrule
  \end{tabular}
\end{table*}

\subsubsection{Performance with missing views.}
HMDR achieves the highest Avg$_{\mathrm{uni}}$ macro F1 among the compared methods, reaching 67.51\% on FineFake and 89.94\% on MDSM (Table~\ref{tab:main}). Compared with Qwen3.5-4B LoRA, which uses the same backbone and fine tuning settings, HMDR improves Avg$_{\mathrm{uni}}$ macro F1 by 6.87 and 17.23 percentage points, respectively. These gains occur in both views. When only text is available, macro F1 increases by 5.66 points on FineFake and 27.67 points on MDSM. When only images are available, the corresponding gains are 8.09 and 6.79 points. FineFake therefore benefits more from improved image classification, while the larger average gain on MDSM mainly comes from improved text classification.

HMDR also outperforms the strongest competing baseline on each dataset. On FineFake, it exceeds the 64.98\% Avg$_{\mathrm{uni}}$ macro F1 of Gemma4-E4B LoRA + MissRAG by 2.53 points. On MDSM, it exceeds the 86.88\% of MIMoE-FND by 3.06 points. For individual views, HMDR ranks first on T in both datasets and on I in MDSM. On FineFake I, its macro F1 of 61.07\% nearly matches the best result of 61.08\% from Gemma4-E4B LoRA + MissRAG. Its average advantage over this baseline on FineFake comes from stronger performance on T while maintaining a similar score on I.

\subsubsection{Performance on complete news.}
On complete news (TI), HMDR reaches 81.73\% macro F1 on FineFake and 89.90\% on MDSM, compared with 81.58\% and 87.32\% for Qwen3.5-4B LoRA. The corresponding gains are 0.15 and 2.58 percentage points. FineFake TI accuracy remains at 81.76\%, while MDSM TI accuracy increases from 87.36\% to 90.01\%. With the same backbone, the improvements under missing views thus accompany unchanged or higher scores on complete news. The improvement on complete news is modest on FineFake and more pronounced on MDSM.

HMDR ranks second in TI macro F1 on both datasets. Gemma4-E4B LoRA achieves the highest scores of 81.85\% on FineFake and 90.14\% on MDSM, exceeding HMDR by 0.12 and 0.24 points, respectively. HMDR also outperforms the three multimodal fake news baselines on TI macro F1. The strongest of these baselines, MIMoE-FND, obtains 80.03\% on FineFake and 89.35\% on MDSM. HMDR therefore remains close to the best detector on complete news while retaining its advantages when a view is missing.

\subsubsection{View imbalance and retrieval.}
The FineFake test set for I contains 1,796 real images and 515 fake images, with real images accounting for 77.72\% of the inputs. MTS, MIMoE-FND, DDU, and AOEPT obtain macro F1 scores between 18\% and 20\%, close to the 18.22\% achieved by predicting every image as fake. Their saved predictions reveal a strong bias toward the fake class. All four methods achieve 100\% recall for fake images, whereas recall for real images ranges from 0\% to 1.11\%. Their high recall for fake images therefore coexists with failure to recognize almost all genuine images. This behavior illustrates the gap between performance on complete news and reliable classification of isolated images. Macro F1 exposes this failure by assigning equal weight to the real and fake classes.

The effect of retrieval varies across datasets and available modalities. On FineFake, adding MissRAG to Qwen LoRA raises macro F1 on I from 52.98\% to 60.57\% but lowers the score on T from 68.28\% to 66.30\%. On MDSM, adding MissRAG lowers both T and I scores for both backbones. Avg$_{\mathrm{uni}}$ macro F1 falls from 72.71\% to 62.12\% for Qwen and from 84.85\% to 62.81\% for Gemma. These results indicate that the benefit of this retrieval adaptation depends on the dataset and the available modality.

\subsection{Ablation Study (RQ2)}
\label{sec:exp_ablation}

We remove MIL, disable MHR, or shuffle the head-group assignments used by MIL (Table~\ref{tab:ablation}). We also test a missing-view regulation policy on the full model's fixed checkpoint.

\subsubsection{MIL and MHR.}
Removing MIL sets $\lambda=0$ while retaining MHD and MHR, so training uses only the classification loss. Avg$_{\mathrm{uni}}$ macro-F1 falls to 61.51\% on FineFake and 77.00\% on MDSM, decreases of 6.00 and 12.94 percentage points from full HMDR (Table~\ref{tab:ablation}). The largest losses occur on FineFake I, which drops from 61.07\% to 52.28\%, and MDSM T, which drops from 97.26\% to 72.45\%. The other missing view also declines on each dataset, indicating that the benefit of MIL extends to both modalities, although its magnitude differs. Complete-view effects are less consistent: removing MIL raises FineFake TI macro-F1 by 1.38 points but lowers MDSM TI by 0.42 points. MIL therefore improves missing-view performance on both datasets while introducing a complete-view trade-off on FineFake.

Disabling MHR sets all head gains to one throughout training and inference while retaining MIL. Avg$_{\mathrm{uni}}$ macro-F1 decreases by 5.90 points on FineFake and 6.88 points on MDSM, reaching 61.61\% and 83.06\%, respectively. FineFake I falls to 51.76\%, while MDSM T and I fall to 87.58\% and 78.54\%. In contrast, TI macro-F1 increases slightly, by 0.13 and 0.21 points. Thus, slightly better complete-news test scores can coexist with poorer missing-view performance. These results support the usefulness of complete-input regulation for subsequent missing-view evaluation.

\subsubsection{Head assignments used by MIL.}
To examine the head assignments used by MIL, we apply a fixed permutation with no fixed points within each layer. Each head receives another head's group-weight vector for computing the auxiliary representations, while MHR retains its original weights. This preserves the distribution of group weights within a layer but changes how head outputs are combined for the MIL loss. On MDSM, Avg$_{\mathrm{uni}}$ macro-F1 drops from 89.94\% to 48.15\%, with T falling from 97.26\% to 19.47\% and I from 82.61\% to 76.83\%. The average score is also well below the 77.00\% obtained without MIL. For this dataset and permutation, applying MIL with shuffled assignments is therefore more harmful than omitting the regularizer.

FineFake shows a smaller effect. Shuffling reduces T macro-F1 from 73.94\% to 73.42\% and I from 61.07\% to 58.98\%, leaving Avg$_{\mathrm{uni}}$ at 66.20\%. Meanwhile, average accuracy increases from 74.88\% to 75.43\%, so the shuffled variant does not underperform on every metric. These results show that the assignment of weights to heads matters for the auxiliary objective, with a much stronger effect on MDSM in the tested runs.

\subsubsection{Native head gains on missing views.}
This variant tests whether missing inputs should retain native head gains or receive modality-dependent scaling. Using the same trained HMDR checkpoint, it scales each head in proportion to its weight for the available modality and normalizes the gains to have a mean of one. Complete inputs retain the original MHR policy, so TI predictions are unchanged. Applying the alternative policy lowers Avg$_{\mathrm{uni}}$ macro-F1 to 63.39\% on FineFake and 88.14\% on MDSM, decreases of 4.12 and 1.80 points, respectively.

The decline is concentrated on image-only inputs. FineFake T improves slightly from 73.94\% to 74.10\%, but I falls from 61.07\% to 52.67\%; on MDSM, T changes from 97.26\% to 96.96\% and I from 82.61\% to 79.32\%. Thus, scaling heads according to the available modality does not improve the two missing-input settings consistently. FineFake average accuracy nevertheless rises from 74.88\% to 77.23\%, while average macro-F1 falls, showing that the two metrics favor different prediction trade-offs on this imbalanced evaluation set. The same-checkpoint results favor keeping native head gains over the tested scaling policy, particularly for image-only classification.

\begin{table}[t]
  \centering
  \caption{Sensitivity to the MIL weight $\lambda$: test macro-F1 (\%) at seed 42. Both datasets use the common default $0.01$; $\lambda=0$ is HMDR without MIL.}
  \label{tab:lambda}
  \small
  \setlength{\tabcolsep}{5pt}
  \begin{tabular}{l cc cc}
    \toprule
    \multirow{2}{*}{$\lambda$} & \multicolumn{2}{c}{FineFake} & \multicolumn{2}{c}{MDSM} \\
    \cmidrule(lr){2-3}\cmidrule(lr){4-5}
    & TI & Avg$_{\mathrm{uni}}$ & TI & Avg$_{\mathrm{uni}}$ \\
    \midrule
    0     & 83.11 & 61.51 & 89.48 & 77.00 \\
    0.001 & 82.46 & 64.05 & 89.56 & 81.40 \\
    0.01  & 81.73 & \textbf{67.51} & 89.90 & \textbf{89.94} \\
    0.1   & 82.05 & 60.06 & 86.05 & 73.59 \\
    \bottomrule
  \end{tabular}
\end{table}
\subsection{Genuine Content Remaining in Fake News (RQ3)}
\label{sec:exp_cohort}

We partition test items by their news label and remaining-view label, requiring both TI and the remaining view to be valid (Table~\ref{tab:cohort}). Under the derived label mapping, this gives Real$\rightarrow$Real, Fake$\rightarrow$Real, and Fake$\rightarrow$Fake cohorts. These are ground-truth label transitions, not changes in a model's predictions.

\subsubsection{Recognizing genuine remaining content.}
For Fake$\rightarrow$Real, Qwen LoRA has accuracies of 20.68\% and 24.75\% on FineFake T and I, and 61.84\% and 12.81\% on MDSM T and I. HMDR improves all four, reaching 44.07\%, 65.42\%, 99.20\%, and 61.94\%, respectively. It also improves every Real$\rightarrow$Real cohort. These results show fewer false alarms on genuine remaining content, including content originating from fake news.

Comparing HMDR with its no-MIL variant further shows that MIL's contribution varies across cohorts. On FineFake I, Fake$\rightarrow$Real accuracy rises from 29.15\% to 65.42\%; on MDSM T, it rises from 68.30\% to 99.20\%. On MDSM I, the no-MIL variant already reaches 56.85\%, versus 61.94\% for full HMDR.

\subsubsection{Retaining detection of fabricated content.}
The improvements are not uniform across classes. On FineFake I, Fake$\rightarrow$Fake recall falls from 71.46\% with LoRA to 40.78\% with HMDR, while genuine-image recognition improves. On the other three views, the corresponding recall changes range from $-4.08$ to $+1.16$ points. HMDR improves overall macro-F1 on all four missing views, but the FineFake image-only result reveals a substantial trade-off between fewer false alarms and detecting fabricated images.

\subsection{Representation Responses to Modality Replacement (RQ4)}
\label{sec:exp_decouple}

We compare HMDR and its no-MIL variant on 24 fixed pairs of complete validation items per dataset. These inputs receive no gradient updates, although the validation split is also used for checkpoint selection. Each pair has compatible token positions, padding masks, and image grids. For items $(a,b)$, we construct $(t_a,v_a)$, $(t_a,v_b)$, $(t_b,v_a)$, and $(t_b,v_b)$. For each modality group and layer, we measure the mean squared change in its representation when one modality is replaced, average over the two values of the fixed modality, and normalize by mean representation energy across the four inputs. This yields responses to the group's own modality and to the other modality.

With MIL, the mean normalized response to the other modality is lower in 7/8 text-group layers and 8/8 image-group layers on FineFake, and in 6/8 and 7/8 layers on MDSM. The measurements therefore support reduced cross-modal sensitivity on these validation pairs

\subsection{Sensitivity}
\label{sec:exp_cost}

Table~\ref{tab:lambda} reports sensitivity around the common default $\lambda=0.01$, retained for both datasets. Among the tested values, $0.01$ gives the highest Avg$_{\mathrm{uni}}$ macro-F1 on both datasets. Raising it to $0.1$ lowers missing-view performance below the no-MIL control and also reduces MDSM TI macro-F1 to 86.05\%. These runs favor a moderate MIL weight within the tested range.

\section{Conclusion}
\label{sec:conclusion}
HMDR detects multimodal fake news with missing modalities, where remaining content may differ in truth label from complete news. It combines attention based head grouping, head regulation, and modality invariance, using only complete news labels and no trainable parameters beyond LoRA. On FineFake and MDSM, it achieves the best average macro F1 under missing modalities while preserving complete news performance.

\bibliographystyle{ACM-Reference-Format}
\bibliography{sample-base}


\begin{thebibliography}{57}


\ifx \showCODEN    \undefined \def \showCODEN     #1{\unskip}     \fi
\ifx \showISBNx    \undefined \def \showISBNx     #1{\unskip}     \fi
\ifx \showISBNxiii \undefined \def \showISBNxiii  #1{\unskip}     \fi
\ifx \showISSN     \undefined \def \showISSN      #1{\unskip}     \fi
\ifx \showLCCN     \undefined \def \showLCCN      #1{\unskip}     \fi
\ifx \shownote     \undefined \def \shownote      #1{#1}          \fi
\ifx \showarticletitle \undefined \def \showarticletitle #1{#1}   \fi
\ifx \showURL      \undefined \def \showURL       {\relax}        \fi
\providecommand\bibfield[2]{#2}
\providecommand\bibinfo[2]{#2}
\providecommand\natexlab[1]{#1}
\providecommand\showeprint[2][]{arXiv:#2}

\bibitem[Bi et~al\mbox{.}(2025)]%
        {bi2025unveiling}
\bibfield{author}{\bibinfo{person}{Jing Bi}, \bibinfo{person}{Junjia Guo},
  \bibinfo{person}{Yunlong Tang}, \bibinfo{person}{Lianggong~Bruce Wen},
  \bibinfo{person}{Zhang Liu}, \bibinfo{person}{Bingjie Wang}, {and}
  \bibinfo{person}{Chenliang Xu}.} \bibinfo{year}{2025}\natexlab{}.
\newblock \showarticletitle{Unveiling visual perception in language models: An
  attention head analysis approach}. In \bibinfo{booktitle}{\emph{Proceedings
  of the Computer Vision and Pattern Recognition Conference}}.
  \bibinfo{pages}{4135--4144}.
\newblock


\bibitem[Chen et~al\mbox{.}(2025)]%
        {chen2025trisprompt}
\bibfield{author}{\bibinfo{person}{Jiajun Chen}, \bibinfo{person}{Yangyang Wu},
  \bibinfo{person}{Xiaoye Miao}, \bibinfo{person}{Mengying Zhu}, {and}
  \bibinfo{person}{Meng Xi}.} \bibinfo{year}{2025}\natexlab{}.
\newblock \showarticletitle{Trisprompt: A hierarchical soft prompt model for
  multimodal rumor detection with incomplete modalities}.
\newblock \bibinfo{journal}{\emph{arXiv preprint arXiv:2509.19352}}
  (\bibinfo{year}{2025}).
\newblock


\bibitem[Chen et~al\mbox{.}(2026)]%
        {chen2026ddu}
\bibfield{author}{\bibinfo{person}{Xuran Chen}, \bibinfo{person}{Xianglin Wei},
  {and} \bibinfo{person}{Yang Yang}.} \bibinfo{year}{2026}\natexlab{}.
\newblock \showarticletitle{{DDU}: Decoupling Dual-Level Uncertainties for
  Reliable Multimodal Fake News Detection}. In
  \bibinfo{booktitle}{\emph{Proceedings of the 34th ACM International
  Conference on Multimedia}}.
\newblock
\href{https://doi.org/10.1145/3767308.3835912}{doi:\nolinkurl{10.1145/3767308.3835912}}


\bibitem[Chen et~al\mbox{.}(2022a)]%
        {chen2022cross}
\bibfield{author}{\bibinfo{person}{Yixuan Chen}, \bibinfo{person}{Dongsheng
  Li}, \bibinfo{person}{Peng Zhang}, \bibinfo{person}{Jie Sui},
  \bibinfo{person}{Qin Lv}, \bibinfo{person}{Lu Tun}, {and} \bibinfo{person}{Li
  Shang}.} \bibinfo{year}{2022}\natexlab{a}.
\newblock \showarticletitle{Cross-modal ambiguity learning for multimodal fake
  news detection}. In \bibinfo{booktitle}{\emph{Proceedings of the ACM web
  conference 2022}}. \bibinfo{pages}{2897--2905}.
\newblock


\bibitem[Chen et~al\mbox{.}(2022b)]%
        {chen2022multimodal}
\bibfield{author}{\bibinfo{person}{Yi-Ting Chen}, \bibinfo{person}{Jinghao
  Shi}, \bibinfo{person}{Zelin Ye}, \bibinfo{person}{Christoph Mertz},
  \bibinfo{person}{Deva Ramanan}, {and} \bibinfo{person}{Shu Kong}.}
  \bibinfo{year}{2022}\natexlab{b}.
\newblock \showarticletitle{Multimodal object detection via probabilistic
  ensembling}. In \bibinfo{booktitle}{\emph{European Conference on Computer
  Vision}}. Springer, \bibinfo{pages}{139--158}.
\newblock


\bibitem[Fisher et~al\mbox{.}(2016)]%
        {fisher2016pizzagate}
\bibfield{author}{\bibinfo{person}{Marc Fisher}, \bibinfo{person}{John~Woodrow
  Cox}, {and} \bibinfo{person}{Peter Hermann}.}
  \bibinfo{year}{2016}\natexlab{}.
\newblock \showarticletitle{Pizzagate: From rumor, to hashtag, to gunfire in
  DC}.
\newblock \bibinfo{journal}{\emph{Washington Post}}  \bibinfo{volume}{6}
  (\bibinfo{year}{2016}), \bibinfo{pages}{8410--8415}.
\newblock


\bibitem[Fu and Liu(2023)]%
        {fu2023multimodal}
\bibfield{author}{\bibinfo{person}{Lifang Fu} {and} \bibinfo{person}{Shuai
  Liu}.} \bibinfo{year}{2023}\natexlab{}.
\newblock \showarticletitle{Multimodal fake news detection incorporating
  external knowledge and user interaction feature}.
\newblock \bibinfo{journal}{\emph{Advances in Multimedia}}
  \bibinfo{volume}{2023}, \bibinfo{number}{1} (\bibinfo{year}{2023}),
  \bibinfo{pages}{8836476}.
\newblock


\bibitem[Ge et~al\mbox{.}(2023)]%
        {ge2023metabev}
\bibfield{author}{\bibinfo{person}{Chongjian Ge}, \bibinfo{person}{Junsong
  Chen}, \bibinfo{person}{Enze Xie}, \bibinfo{person}{Zhongdao Wang},
  \bibinfo{person}{Lanqing Hong}, \bibinfo{person}{Huchuan Lu},
  \bibinfo{person}{Zhenguo Li}, {and} \bibinfo{person}{Ping Luo}.}
  \bibinfo{year}{2023}\natexlab{}.
\newblock \showarticletitle{Metabev: Solving sensor failures for 3d detection
  and map segmentation}. In \bibinfo{booktitle}{\emph{Proceedings of the
  IEEE/CVF International Conference on Computer Vision}}.
  \bibinfo{pages}{8721--8731}.
\newblock


\bibitem[{Gemma Team} et~al\mbox{.}(2026)]%
        {gemmateam2026gemma4}
\bibfield{author}{\bibinfo{person}{{Gemma Team}}, \bibinfo{person}{Sherif~El
  Abd}, \bibinfo{person}{Vaibhav Aggarwal}, \bibinfo{person}{Robin Algayres},
  \bibinfo{person}{Alek Andreev}, \bibinfo{person}{Olivier Bachem},
  \bibinfo{person}{Ian Ballantyne}, \bibinfo{person}{Cormac Brick},
  \bibinfo{person}{Victor Cărbune}, \bibinfo{person}{Michelle Casbon},
  {et~al\mbox{.}}} \bibinfo{year}{2026}\natexlab{}.
\newblock \bibinfo{title}{{Gemma 4} Technical Report}.
\newblock
\showeprint[arxiv]{2607.02770}~[cs.CL]
\urldef\tempurl%
\url{https://arxiv.org/abs/2607.02770}
\showURL{%
\tempurl}


\bibitem[Guo et~al\mbox{.}(2025)]%
        {guo2025each}
\bibfield{author}{\bibinfo{person}{Hao Guo}, \bibinfo{person}{Zihan Ma},
  \bibinfo{person}{Zhi Zeng}, \bibinfo{person}{Minnan Luo},
  \bibinfo{person}{Weixin Zeng}, \bibinfo{person}{Jiuyang Tang}, {and}
  \bibinfo{person}{Xiang Zhao}.} \bibinfo{year}{2025}\natexlab{}.
\newblock \showarticletitle{Each fake news is fake in its own way: An
  attribution multi-granularity benchmark for multimodal fake news detection}.
  In \bibinfo{booktitle}{\emph{Proceedings of the AAAI conference on artificial
  intelligence}}, Vol.~\bibinfo{volume}{39}. \bibinfo{pages}{228--236}.
\newblock


\bibitem[Guo et~al\mbox{.}(2023)]%
        {guo2023two}
\bibfield{author}{\bibinfo{person}{Ying Guo}, \bibinfo{person}{Hong Ge}, {and}
  \bibinfo{person}{Jinhong Li}.} \bibinfo{year}{2023}\natexlab{}.
\newblock \showarticletitle{A two-branch multimodal fake news detection model
  based on multimodal bilinear pooling and attention mechanism}.
\newblock \bibinfo{journal}{\emph{Frontiers in Computer Science}}
  \bibinfo{volume}{5} (\bibinfo{year}{2023}), \bibinfo{pages}{1159063}.
\newblock


\bibitem[He et~al\mbox{.}(2025)]%
        {he2025cracking}
\bibfield{author}{\bibinfo{person}{Jinghan He}, \bibinfo{person}{Kuan Zhu},
  \bibinfo{person}{Haiyun Guo}, \bibinfo{person}{Junfeng Fang},
  \bibinfo{person}{Zhenglin Hua}, \bibinfo{person}{Yuheng Jia},
  \bibinfo{person}{Ming Tang}, \bibinfo{person}{Tat-Seng Chua}, {and}
  \bibinfo{person}{Jinqiao Wang}.} \bibinfo{year}{2025}\natexlab{}.
\newblock \showarticletitle{Cracking the code of hallucination in lvlms with
  vision-aware head divergence}. In \bibinfo{booktitle}{\emph{Proceedings of
  the 63rd Annual Meeting of the Association for Computational Linguistics
  (Volume 1: Long Papers)}}. \bibinfo{pages}{3488--3501}.
\newblock


\bibitem[Hoffman et~al\mbox{.}(2016)]%
        {hoffman2016learning}
\bibfield{author}{\bibinfo{person}{Judy Hoffman}, \bibinfo{person}{Saurabh
  Gupta}, {and} \bibinfo{person}{Trevor Darrell}.}
  \bibinfo{year}{2016}\natexlab{}.
\newblock \showarticletitle{Learning with side information through modality
  hallucination}. In \bibinfo{booktitle}{\emph{Proceedings of the IEEE
  conference on computer vision and pattern recognition}}.
  \bibinfo{pages}{826--834}.
\newblock


\bibitem[Hu et~al\mbox{.}(2022)]%
        {hu2022lora}
\bibfield{author}{\bibinfo{person}{Edward~J. Hu}, \bibinfo{person}{Yelong
  Shen}, \bibinfo{person}{Phillip Wallis}, \bibinfo{person}{Zeyuan Allen-Zhu},
  \bibinfo{person}{Yuanzhi Li}, \bibinfo{person}{Shean Wang},
  \bibinfo{person}{Lu Wang}, {and} \bibinfo{person}{Weizhu Chen}.}
  \bibinfo{year}{2022}\natexlab{}.
\newblock \showarticletitle{{LoRA}: Low-Rank Adaptation of Large Language
  Models}. In \bibinfo{booktitle}{\emph{International Conference on Learning
  Representations}}.
\newblock
\urldef\tempurl%
\url{https://openreview.net/forum?id=nZeVKeeFYf9}
\showURL{%
\tempurl}


\bibitem[Hua et~al\mbox{.}(2023)]%
        {hua2023multimodal}
\bibfield{author}{\bibinfo{person}{Jiaheng Hua}, \bibinfo{person}{Xiaodong
  Cui}, \bibinfo{person}{Xianghua Li}, \bibinfo{person}{Keke Tang}, {and}
  \bibinfo{person}{Peican Zhu}.} \bibinfo{year}{2023}\natexlab{}.
\newblock \showarticletitle{Multimodal fake news detection through data
  augmentation-based contrastive learning}.
\newblock \bibinfo{journal}{\emph{Applied Soft Computing}}
  \bibinfo{volume}{136} (\bibinfo{year}{2023}), \bibinfo{pages}{110125}.
\newblock


\bibitem[Jin et~al\mbox{.}(2024)]%
        {jin2024fake}
\bibfield{author}{\bibinfo{person}{Ruihan Jin}, \bibinfo{person}{Ruibo Fu},
  \bibinfo{person}{Zhengqi Wen}, \bibinfo{person}{Shuai Zhang},
  \bibinfo{person}{Yukun Liu}, {and} \bibinfo{person}{Jianhua Tao}.}
  \bibinfo{year}{2024}\natexlab{}.
\newblock \showarticletitle{Fake news detection and manipulation reasoning via
  large vision-language models}.
\newblock \bibinfo{journal}{\emph{arXiv preprint arXiv:2407.02042}}
  (\bibinfo{year}{2024}).
\newblock


\bibitem[Jin et~al\mbox{.}(2017)]%
        {jin2017multimodal}
\bibfield{author}{\bibinfo{person}{Zhiwei Jin}, \bibinfo{person}{Juan Cao},
  \bibinfo{person}{Han Guo}, \bibinfo{person}{Yongdong Zhang}, {and}
  \bibinfo{person}{Jiebo Luo}.} \bibinfo{year}{2017}\natexlab{}.
\newblock \showarticletitle{Multimodal fusion with recurrent neural networks
  for rumor detection on microblogs}. In \bibinfo{booktitle}{\emph{Proceedings
  of the 25th ACM international conference on Multimedia}}.
  \bibinfo{pages}{795--816}.
\newblock


\bibitem[Kang et~al\mbox{.}(2025)]%
        {kang2025see}
\bibfield{author}{\bibinfo{person}{Seil Kang}, \bibinfo{person}{Jinyeong Kim},
  \bibinfo{person}{Junhyeok Kim}, {and} \bibinfo{person}{Seong~Jae Hwang}.}
  \bibinfo{year}{2025}\natexlab{}.
\newblock \showarticletitle{See what you are told: Visual attention sink in
  large multimodal models}. In \bibinfo{booktitle}{\emph{International
  Conference on Learning Representations}}, Vol.~\bibinfo{volume}{2025}.
  \bibinfo{pages}{87676--87703}.
\newblock


\bibitem[Ke et~al\mbox{.}(2025)]%
        {ke2025knowledge}
\bibfield{author}{\bibinfo{person}{Guanzhou Ke}, \bibinfo{person}{Shengfeng
  He}, \bibinfo{person}{Xiaoli Wang}, \bibinfo{person}{Bo Wang},
  \bibinfo{person}{Guoqing Chao}, \bibinfo{person}{Yuanyang Zhang},
  \bibinfo{person}{Yi Xie}, {and} \bibinfo{person}{Hexing Su}.}
  \bibinfo{year}{2025}\natexlab{}.
\newblock \showarticletitle{Knowledge bridger: Towards training-free missing
  modality completion}. In \bibinfo{booktitle}{\emph{Proceedings of the
  Computer Vision and Pattern Recognition Conference}}.
  \bibinfo{pages}{25864--25873}.
\newblock


\bibitem[Lang et~al\mbox{.}(2026)]%
        {lang2026aoept}
\bibfield{author}{\bibinfo{person}{Jian Lang}, \bibinfo{person}{Rongpei Hong},
  \bibinfo{person}{Ting Zhong}, {and} \bibinfo{person}{Fan Zhou}.}
  \bibinfo{year}{2026}\natexlab{}.
\newblock \showarticletitle{{AOEPT}: Breaking the Implicit Modality-Reduction
  Bottleneck in Modality-Missing Prompt Tuning}. In
  \bibinfo{booktitle}{\emph{Proceedings of the 43rd International Conference on
  Machine Learning}} \emph{(\bibinfo{series}{Proceedings of Machine Learning
  Research}, Vol.~\bibinfo{volume}{306})}. \bibinfo{publisher}{PMLR},
  \bibinfo{pages}{62583--62602}.
\newblock
\urldef\tempurl%
\url{https://proceedings.mlr.press/v306/lang26a.html}
\showURL{%
\tempurl}


\bibitem[Lee et~al\mbox{.}(2023a)]%
        {lee2023learning}
\bibfield{author}{\bibinfo{person}{Kwanhyung Lee}, \bibinfo{person}{Soojeong
  Lee}, \bibinfo{person}{Sangchul Hahn}, \bibinfo{person}{Heejung Hyun},
  \bibinfo{person}{Edward Choi}, \bibinfo{person}{Byungeun Ahn}, {and}
  \bibinfo{person}{Joohyung Lee}.} \bibinfo{year}{2023}\natexlab{a}.
\newblock \showarticletitle{Learning missing modal electronic health records
  with unified multi-modal data embedding and modality-aware attention}. In
  \bibinfo{booktitle}{\emph{Machine Learning for Healthcare Conference}}. PMLR,
  \bibinfo{pages}{423--442}.
\newblock


\bibitem[Lee et~al\mbox{.}(2023b)]%
        {lee2023multimodal}
\bibfield{author}{\bibinfo{person}{Yi-Lun Lee}, \bibinfo{person}{Yi-Hsuan
  Tsai}, \bibinfo{person}{Wei-Chen Chiu}, {and} \bibinfo{person}{Chen-Yu Lee}.}
  \bibinfo{year}{2023}\natexlab{b}.
\newblock \showarticletitle{Multimodal prompting with missing modalities for
  visual recognition}. In \bibinfo{booktitle}{\emph{2023 IEEE/CVF Conference on
  Computer Vision and Pattern Recognition (CVPR)}}. IEEE,
  \bibinfo{pages}{14943--14952}.
\newblock


\bibitem[Li et~al\mbox{.}(2025)]%
        {li2025entity}
\bibfield{author}{\bibinfo{person}{Guoyi Li}, \bibinfo{person}{Die Hu},
  \bibinfo{person}{Xiaomeng Fu}, \bibinfo{person}{Qirui Tang},
  \bibinfo{person}{Yulei Wu}, \bibinfo{person}{Xiaodan Zhang}, {and}
  \bibinfo{person}{Honglei Lyu}.} \bibinfo{year}{2025}\natexlab{}.
\newblock \showarticletitle{Entity Graph Alignment and Visual Reasoning for
  Multimodal Fake News Detection}. In \bibinfo{booktitle}{\emph{Proceedings of
  the 33rd ACM International Conference on Multimedia}}.
  \bibinfo{pages}{2486--2495}.
\newblock


\bibitem[Li et~al\mbox{.}(2023)]%
        {li2023inference}
\bibfield{author}{\bibinfo{person}{Kenneth Li}, \bibinfo{person}{Oam Patel},
  \bibinfo{person}{Fernanda Vi{\'e}gas}, \bibinfo{person}{Hanspeter Pfister},
  {and} \bibinfo{person}{Martin Wattenberg}.} \bibinfo{year}{2023}\natexlab{}.
\newblock \showarticletitle{Inference-time intervention: Eliciting truthful
  answers from a language model}.
\newblock \bibinfo{journal}{\emph{Advances in neural information processing
  systems}}  \bibinfo{volume}{36} (\bibinfo{year}{2023}),
  \bibinfo{pages}{41451--41530}.
\newblock


\bibitem[Liang et~al\mbox{.}(2019)]%
        {liang2019learning}
\bibfield{author}{\bibinfo{person}{Paul~Pu Liang}, \bibinfo{person}{Zhun Liu},
  \bibinfo{person}{Yao-Hung~Hubert Tsai}, \bibinfo{person}{Qibin Zhao},
  \bibinfo{person}{Ruslan Salakhutdinov}, {and} \bibinfo{person}{Louis-Philippe
  Morency}.} \bibinfo{year}{2019}\natexlab{}.
\newblock \showarticletitle{Learning representations from imperfect time series
  data via tensor rank regularization}. In
  \bibinfo{booktitle}{\emph{Proceedings of the 57th Annual Meeting of the
  Association for Computational Linguistics}}. \bibinfo{pages}{1569--1576}.
\newblock


\bibitem[Liu et~al\mbox{.}(2025)]%
        {liu2025modality}
\bibfield{author}{\bibinfo{person}{Yifan Liu}, \bibinfo{person}{Yaokun Liu},
  \bibinfo{person}{Zelin Li}, \bibinfo{person}{Ruichen Yao},
  \bibinfo{person}{Yang Zhang}, {and} \bibinfo{person}{Dong Wang}.}
  \bibinfo{year}{2025}\natexlab{}.
\newblock \showarticletitle{Modality Interactive Mixture-of-Experts for Fake
  News Detection}. In \bibinfo{booktitle}{\emph{Proceedings of the ACM on Web
  Conference 2025}}. \bibinfo{pages}{5139--5150}.
\newblock
\href{https://doi.org/10.1145/3696410.3714522}{doi:\nolinkurl{10.1145/3696410.3714522}}


\bibitem[Loshchilov and Hutter(2019)]%
        {loshchilov2019decoupled}
\bibfield{author}{\bibinfo{person}{Ilya Loshchilov} {and}
  \bibinfo{person}{Frank Hutter}.} \bibinfo{year}{2019}\natexlab{}.
\newblock \showarticletitle{Decoupled Weight Decay Regularization}. In
  \bibinfo{booktitle}{\emph{International Conference on Learning
  Representations}}.
\newblock
\urldef\tempurl%
\url{https://openreview.net/forum?id=Bkg6RiCqY7}
\showURL{%
\tempurl}


\bibitem[Lu et~al\mbox{.}(2026)]%
        {lu2026dpl}
\bibfield{author}{\bibinfo{person}{Jueqing Lu}, \bibinfo{person}{Yuanyuan Qi},
  \bibinfo{person}{Xiaohao Yang}, \bibinfo{person}{Shuaicheng Niu},
  \bibinfo{person}{Fucai Ke}, \bibinfo{person}{Shujie Zhou},
  \bibinfo{person}{Wei Tan}, \bibinfo{person}{Jionghao Lin},
  \bibinfo{person}{Wray Buntine}, \bibinfo{person}{Hamid Rezatofighi}, {and}
  \bibinfo{person}{Lan Du}.} \bibinfo{year}{2026}\natexlab{}.
\newblock \showarticletitle{{DPL}: Decoupled Prototype Learning for Enhancing
  Robustness of Vision-Language Transformers to Missing Modalities}. In
  \bibinfo{booktitle}{\emph{Proceedings of the IEEE/CVF Conference on Computer
  Vision and Pattern Recognition}}. \bibinfo{pages}{39219--39229}.
\newblock
\urldef\tempurl%
\url{https://openaccess.thecvf.com/content/CVPR2026/html/Lu_DPL_Decoupled_Prototype_Learning_for_Enhancing_Robustness_of_Vision-Language_Transformers_CVPR_2026_paper.html}
\showURL{%
\tempurl}


\bibitem[Ma et~al\mbox{.}(2022)]%
        {ma2022multimodal}
\bibfield{author}{\bibinfo{person}{Mengmeng Ma}, \bibinfo{person}{Jian Ren},
  \bibinfo{person}{Long Zhao}, \bibinfo{person}{Davide Testuggine}, {and}
  \bibinfo{person}{Xi Peng}.} \bibinfo{year}{2022}\natexlab{}.
\newblock \showarticletitle{Are multimodal transformers robust to missing
  modality?}. In \bibinfo{booktitle}{\emph{2022 IEEE/CVF conference on computer
  vision and pattern recognition (CVPR)}}. IEEE, \bibinfo{pages}{18156--18165}.
\newblock


\bibitem[Naeem and Bhatti(2020)]%
        {naeem2020covid}
\bibfield{author}{\bibinfo{person}{Salman~Bin Naeem} {and}
  \bibinfo{person}{Rubina Bhatti}.} \bibinfo{year}{2020}\natexlab{}.
\newblock \showarticletitle{The Covid-19 ‘infodemic’: a new front for
  information professionals}.
\newblock \bibinfo{journal}{\emph{Health Information \& Libraries Journal}}
  \bibinfo{volume}{37}, \bibinfo{number}{3} (\bibinfo{year}{2020}),
  \bibinfo{pages}{233--239}.
\newblock


\bibitem[Ognyanova et~al\mbox{.}(2020)]%
        {ognyanova2020misinformation}
\bibfield{author}{\bibinfo{person}{Katherine Ognyanova}, \bibinfo{person}{David
  Lazer}, \bibinfo{person}{Ronald~E Robertson}, {and} \bibinfo{person}{Christo
  Wilson}.} \bibinfo{year}{2020}\natexlab{}.
\newblock \showarticletitle{Misinformation in action: Fake news exposure is
  linked to lower trust in media, higher trust in government when your side is
  in power}.
\newblock \bibinfo{journal}{\emph{Harvard Kennedy School Misinformation
  Review}} (\bibinfo{year}{2020}).
\newblock


\bibitem[Pipoli et~al\mbox{.}(2025)]%
        {pipoli2025missrag}
\bibfield{author}{\bibinfo{person}{Vittorio Pipoli}, \bibinfo{person}{Alessia
  Saporita}, \bibinfo{person}{Federico Bolelli}, \bibinfo{person}{Marcella
  Cornia}, \bibinfo{person}{Lorenzo Baraldi}, \bibinfo{person}{Costantino
  Grana}, \bibinfo{person}{Rita Cucchiara}, {and} \bibinfo{person}{Elisa
  Ficarra}.} \bibinfo{year}{2025}\natexlab{}.
\newblock \showarticletitle{{MissRAG}: Addressing the Missing Modality
  Challenge in Multimodal Large Language Models}. In
  \bibinfo{booktitle}{\emph{Proceedings of the IEEE/CVF International
  Conference on Computer Vision}}. \bibinfo{pages}{3215--3224}.
\newblock
\urldef\tempurl%
\url{https://openaccess.thecvf.com/content/ICCV2025/html/Pipoli_MissRAG_Addressing_the_Missing_Modality_Challenge_in_Multimodal_Large_Language_ICCV_2025_paper.html}
\showURL{%
\tempurl}


\bibitem[{Qwen Team}(2026)]%
        {qwen2026qwen35}
\bibfield{author}{\bibinfo{person}{{Qwen Team}}.}
  \bibinfo{year}{2026}\natexlab{}.
\newblock \bibinfo{title}{{Qwen3.5}: Towards Native Multimodal Agents}.
\newblock
\urldef\tempurl%
\url{https://qwen.ai/blog?id=qwen3.5}
\showURL{%
\tempurl}


\bibitem[Segura-Bedmar and Alonso-Bartolome(2022)]%
        {segura2022multimodal}
\bibfield{author}{\bibinfo{person}{Isabel Segura-Bedmar} {and}
  \bibinfo{person}{Santiago Alonso-Bartolome}.}
  \bibinfo{year}{2022}\natexlab{}.
\newblock \showarticletitle{Multimodal fake news detection}.
\newblock \bibinfo{journal}{\emph{Information}} \bibinfo{volume}{13},
  \bibinfo{number}{6} (\bibinfo{year}{2022}), \bibinfo{pages}{284}.
\newblock


\bibitem[Shang et~al\mbox{.}(2025)]%
        {shang2025semantic}
\bibfield{author}{\bibinfo{person}{Wenqian Shang}, \bibinfo{person}{Kang Song},
  \bibinfo{person}{Jialing Ji}, \bibinfo{person}{Tong Yi},
  \bibinfo{person}{Jiajun Cai}, {and} \bibinfo{person}{Xianxian Li}.}
  \bibinfo{year}{2025}\natexlab{}.
\newblock \showarticletitle{Semantic space aligned multimodal fake news
  detection}.
\newblock \bibinfo{journal}{\emph{Information Fusion}} (\bibinfo{year}{2025}),
  \bibinfo{pages}{103469}.
\newblock


\bibitem[Shao et~al\mbox{.}(2023)]%
        {shao2023detecting}
\bibfield{author}{\bibinfo{person}{Rui Shao}, \bibinfo{person}{Tianxing Wu},
  {and} \bibinfo{person}{Ziwei Liu}.} \bibinfo{year}{2023}\natexlab{}.
\newblock \showarticletitle{Detecting and grounding multi-modal media
  manipulation}. In \bibinfo{booktitle}{\emph{Proceedings of the IEEE/CVF
  Conference on Computer Vision and Pattern Recognition}}.
  \bibinfo{pages}{6904--6913}.
\newblock


\bibitem[Singhal et~al\mbox{.}(2019)]%
        {singhal2019spotfake}
\bibfield{author}{\bibinfo{person}{Shivangi Singhal},
  \bibinfo{person}{Rajiv~Ratn Shah}, \bibinfo{person}{Tanmoy Chakraborty},
  \bibinfo{person}{Ponnurangam Kumaraguru}, {and} \bibinfo{person}{Shin'ichi
  Satoh}.} \bibinfo{year}{2019}\natexlab{}.
\newblock \showarticletitle{Spotfake: A multi-modal framework for fake news
  detection}. In \bibinfo{booktitle}{\emph{2019 IEEE fifth international
  conference on multimedia big data (BigMM)}}. IEEE, \bibinfo{pages}{39--47}.
\newblock


\bibitem[Sun et~al\mbox{.}(2025)]%
        {sun2025multimodal}
\bibfield{author}{\bibinfo{person}{Jiahao Sun}, \bibinfo{person}{Chen Chen},
  \bibinfo{person}{Chunyan Hou}, \bibinfo{person}{Yike Wu}, {and}
  \bibinfo{person}{Xiaojie Yuan}.} \bibinfo{year}{2025}\natexlab{}.
\newblock \showarticletitle{Multimodal Taylor Series Network for Misinformation
  Detection}. In \bibinfo{booktitle}{\emph{Proceedings of the ACM on Web
  Conference 2025}}. \bibinfo{pages}{2540--2548}.
\newblock
\href{https://doi.org/10.1145/3696410.3714719}{doi:\nolinkurl{10.1145/3696410.3714719}}


\bibitem[Tuan and Minh(2021)]%
        {tuan2021multimodal}
\bibfield{author}{\bibinfo{person}{Nguyen Manh~Duc Tuan} {and}
  \bibinfo{person}{Pham Quang~Nhat Minh}.} \bibinfo{year}{2021}\natexlab{}.
\newblock \showarticletitle{Multimodal fusion with BERT and attention mechanism
  for fake news detection}. In \bibinfo{booktitle}{\emph{2021 RIVF
  international conference on computing and communication technologies
  (RIVF)}}. IEEE, \bibinfo{pages}{1--6}.
\newblock


\bibitem[Voita et~al\mbox{.}(2019)]%
        {voita2019analyzing}
\bibfield{author}{\bibinfo{person}{Elena Voita}, \bibinfo{person}{David
  Talbot}, \bibinfo{person}{Fedor Moiseev}, \bibinfo{person}{Rico Sennrich},
  {and} \bibinfo{person}{Ivan Titov}.} \bibinfo{year}{2019}\natexlab{}.
\newblock \showarticletitle{Analyzing multi-head self-attention: Specialized
  heads do the heavy lifting, the rest can be pruned}. In
  \bibinfo{booktitle}{\emph{Proceedings of the 57th annual meeting of the
  association for computational linguistics}}. \bibinfo{pages}{5797--5808}.
\newblock


\bibitem[Wang et~al\mbox{.}(2023b)]%
        {wang2023cross}
\bibfield{author}{\bibinfo{person}{Longzheng Wang}, \bibinfo{person}{Chuang
  Zhang}, \bibinfo{person}{Hongbo Xu}, \bibinfo{person}{Yongxiu Xu},
  \bibinfo{person}{Xiaohan Xu}, {and} \bibinfo{person}{Siqi Wang}.}
  \bibinfo{year}{2023}\natexlab{b}.
\newblock \showarticletitle{Cross-modal contrastive learning for multimodal
  fake news detection}. In \bibinfo{booktitle}{\emph{Proceedings of the 31st
  ACM international conference on multimedia}}. \bibinfo{pages}{5696--5704}.
\newblock


\bibitem[Wang et~al\mbox{.}(2020)]%
        {wang2020icmsc}
\bibfield{author}{\bibinfo{person}{Qianqian Wang}, \bibinfo{person}{Huanhuan
  Lian}, \bibinfo{person}{Gan Sun}, \bibinfo{person}{Quanxue Gao}, {and}
  \bibinfo{person}{Licheng Jiao}.} \bibinfo{year}{2020}\natexlab{}.
\newblock \showarticletitle{iCmSC: Incomplete cross-modal subspace clustering}.
\newblock \bibinfo{journal}{\emph{IEEE Transactions on Image Processing}}
  \bibinfo{volume}{30} (\bibinfo{year}{2020}), \bibinfo{pages}{305--317}.
\newblock


\bibitem[Wang et~al\mbox{.}(2024)]%
        {wang2024fedmmr}
\bibfield{author}{\bibinfo{person}{Shu Wang}, \bibinfo{person}{Zhe Qu},
  \bibinfo{person}{Yuan Liu}, \bibinfo{person}{Shichao Kan},
  \bibinfo{person}{Yixiong Liang}, {and} \bibinfo{person}{Jianxin Wang}.}
  \bibinfo{year}{2024}\natexlab{}.
\newblock \showarticletitle{Fedmmr: Multi-modal federated learning via missing
  modality reconstruction}. In \bibinfo{booktitle}{\emph{2024 IEEE
  International Conference on Multimedia and Expo (ICME)}}. IEEE,
  \bibinfo{pages}{1--6}.
\newblock


\bibitem[Wang et~al\mbox{.}(2023a)]%
        {wang2023distribution}
\bibfield{author}{\bibinfo{person}{Yuanzhi Wang}, \bibinfo{person}{Zhen Cui},
  {and} \bibinfo{person}{Yong Li}.} \bibinfo{year}{2023}\natexlab{a}.
\newblock \showarticletitle{Distribution-consistent modal recovering for
  incomplete multimodal learning}. In \bibinfo{booktitle}{\emph{Proceedings of
  the IEEE/CVF International Conference on Computer Vision}}.
  \bibinfo{pages}{22025--22034}.
\newblock


\bibitem[Wu et~al\mbox{.}(2024)]%
        {wu2024deep}
\bibfield{author}{\bibinfo{person}{Renjie Wu}, \bibinfo{person}{Hu Wang},
  \bibinfo{person}{Hsiang-Ting Chen}, {and} \bibinfo{person}{Gustavo
  Carneiro}.} \bibinfo{year}{2024}\natexlab{}.
\newblock \showarticletitle{Deep multimodal learning with missing modality: A
  survey}.
\newblock \bibinfo{journal}{\emph{arXiv preprint arXiv:2409.07825}}
  (\bibinfo{year}{2024}).
\newblock


\bibitem[Wu et~al\mbox{.}(2021)]%
        {wu2021multimodal}
\bibfield{author}{\bibinfo{person}{Yang Wu}, \bibinfo{person}{Pengwei Zhan},
  \bibinfo{person}{Yunjian Zhang}, \bibinfo{person}{Liming Wang}, {and}
  \bibinfo{person}{Zhen Xu}.} \bibinfo{year}{2021}\natexlab{}.
\newblock \showarticletitle{Multimodal fusion with co-attention networks for
  fake news detection}. In \bibinfo{booktitle}{\emph{Findings of the
  association for computational linguistics: ACL-IJCNLP 2021}}.
  \bibinfo{pages}{2560--2569}.
\newblock


\bibitem[Xu et~al\mbox{.}(2023)]%
        {xu2023leveraging}
\bibfield{author}{\bibinfo{person}{Fan Xu}, \bibinfo{person}{Pinyun Fu},
  \bibinfo{person}{Qi Huang}, \bibinfo{person}{Bowei Zou},
  \bibinfo{person}{AiTi Aw}, {and} \bibinfo{person}{Mingwen Wang}.}
  \bibinfo{year}{2023}\natexlab{}.
\newblock \showarticletitle{Leveraging contrastive learning and knowledge
  distillation for incomplete modality rumor detection}. In
  \bibinfo{booktitle}{\emph{Findings of the Association for Computational
  Linguistics: EMNLP 2023}}. \bibinfo{pages}{13492--13503}.
\newblock


\bibitem[Xu et~al\mbox{.}(2024)]%
        {xu2024leveraging}
\bibfield{author}{\bibinfo{person}{Wenxin Xu}, \bibinfo{person}{Hexin Jiang},
  {and} \bibinfo{person}{Xuefeng Liang}.} \bibinfo{year}{2024}\natexlab{}.
\newblock \showarticletitle{Leveraging knowledge of modality experts for
  incomplete multimodal learning}. In \bibinfo{booktitle}{\emph{Proceedings of
  the 32nd ACM International Conference on Multimedia}}.
  \bibinfo{pages}{438--446}.
\newblock


\bibitem[Xue et~al\mbox{.}(2021)]%
        {xue2021detecting}
\bibfield{author}{\bibinfo{person}{Junxiao Xue}, \bibinfo{person}{Yabo Wang},
  \bibinfo{person}{Yichen Tian}, \bibinfo{person}{Yafei Li},
  \bibinfo{person}{Lei Shi}, {and} \bibinfo{person}{Lin Wei}.}
  \bibinfo{year}{2021}\natexlab{}.
\newblock \showarticletitle{Detecting fake news by exploring the consistency of
  multimodal data}.
\newblock \bibinfo{journal}{\emph{Information Processing \& Management}}
  \bibinfo{volume}{58}, \bibinfo{number}{5} (\bibinfo{year}{2021}),
  \bibinfo{pages}{102610}.
\newblock


\bibitem[Ying et~al\mbox{.}(2021)]%
        {ying2021multi}
\bibfield{author}{\bibinfo{person}{Long Ying}, \bibinfo{person}{Hui Yu},
  \bibinfo{person}{Jinguang Wang}, \bibinfo{person}{Yongze Ji}, {and}
  \bibinfo{person}{Shengsheng Qian}.} \bibinfo{year}{2021}\natexlab{}.
\newblock \showarticletitle{Multi-level multi-modal cross-attention network for
  fake news detection}.
\newblock \bibinfo{journal}{\emph{Ieee Access}}  \bibinfo{volume}{9}
  (\bibinfo{year}{2021}), \bibinfo{pages}{132363--132373}.
\newblock


\bibitem[Zeng et~al\mbox{.}(2024)]%
        {zeng2024missing}
\bibfield{author}{\bibinfo{person}{Zhilin Zeng}, \bibinfo{person}{Zelin Peng},
  \bibinfo{person}{Xiaokang Yang}, {and} \bibinfo{person}{Wei Shen}.}
  \bibinfo{year}{2024}\natexlab{}.
\newblock \showarticletitle{Missing as masking: arbitrary cross-modal feature
  reconstruction for incomplete multimodal brain tumor segmentation}. In
  \bibinfo{booktitle}{\emph{International Conference on Medical Image Computing
  and Computer-Assisted Intervention}}. Springer, \bibinfo{pages}{424--433}.
\newblock


\bibitem[Zeng et~al\mbox{.}(2025)]%
        {zeng2025imol}
\bibfield{author}{\bibinfo{person}{Zhi Zeng}, \bibinfo{person}{Jiaying Wu},
  \bibinfo{person}{Minnan Luo}, \bibinfo{person}{Herun Wan},
  \bibinfo{person}{Xiangzheng Kong}, \bibinfo{person}{Zihan Ma},
  \bibinfo{person}{Guang Dai}, {and} \bibinfo{person}{Qinghua Zheng}.}
  \bibinfo{year}{2025}\natexlab{}.
\newblock \showarticletitle{Imol: Incomplete-modality-tolerant learning for
  multi-domain fake news video detection}. In
  \bibinfo{booktitle}{\emph{Proceedings of the 63rd Annual Meeting of the
  Association for Computational Linguistics (Volume 1: Long Papers)}}.
  \bibinfo{pages}{30921--30933}.
\newblock


\bibitem[Zhan et~al\mbox{.}(2025)]%
        {zhan2025systematic}
\bibfield{author}{\bibinfo{person}{Yifan Zhan}, \bibinfo{person}{Rui Yang},
  \bibinfo{person}{Junxian You}, \bibinfo{person}{Mengjie Huang},
  \bibinfo{person}{Weibo Liu}, {and} \bibinfo{person}{Xiaohui Liu}.}
  \bibinfo{year}{2025}\natexlab{}.
\newblock \showarticletitle{A systematic literature review on incomplete
  multimodal learning: techniques and challenges}.
\newblock \bibinfo{journal}{\emph{Systems Science \& Control Engineering}}
  \bibinfo{volume}{13}, \bibinfo{number}{1} (\bibinfo{year}{2025}),
  \bibinfo{pages}{2467083}.
\newblock


\bibitem[Zhang et~al\mbox{.}(2026b)]%
        {zhang2026locate}
\bibfield{author}{\bibinfo{person}{Hengyuan Zhang}, \bibinfo{person}{Zhihao
  Zhang}, \bibinfo{person}{Mingyang Wang}, \bibinfo{person}{Zunhai Su},
  \bibinfo{person}{Yiwei Wang}, \bibinfo{person}{Qianli Wang},
  \bibinfo{person}{Shuzhou Yuan}, \bibinfo{person}{Ercong Nie},
  \bibinfo{person}{Xufeng Duan}, \bibinfo{person}{Qibo Xue}, {et~al\mbox{.}}}
  \bibinfo{year}{2026}\natexlab{b}.
\newblock \showarticletitle{Locate, Steer, and Improve: A Practical Survey of
  Actionable Mechanistic Interpretability in Large Language Models}.
\newblock \bibinfo{journal}{\emph{arXiv preprint arXiv:2601.14004}}
  (\bibinfo{year}{2026}).
\newblock


\bibitem[Zhang et~al\mbox{.}(2026a)]%
        {zhang2026coherence}
\bibfield{author}{\bibinfo{person}{Yuchen Zhang}, \bibinfo{person}{Yaxiong
  Wang}, \bibinfo{person}{Yujiao Wu}, \bibinfo{person}{Lianwei Wu},
  \bibinfo{person}{Li Zhu}, {and} \bibinfo{person}{Zhedong Zheng}.}
  \bibinfo{year}{2026}\natexlab{a}.
\newblock \showarticletitle{The Coherence Trap: When {MLLM}-Crafted Narratives
  Exploit Manipulated Visual Contexts}. In
  \bibinfo{booktitle}{\emph{Proceedings of the IEEE/CVF Conference on Computer
  Vision and Pattern Recognition}}. \bibinfo{pages}{8760--8769}.
\newblock
\urldef\tempurl%
\url{https://openaccess.thecvf.com/content/CVPR2026/papers/Zhang_The_Coherence_Trap_When_MLLM-Crafted_Narratives_Exploit_Manipulated_Visual_Contexts_CVPR_2026_paper.pdf}
\showURL{%
\tempurl}


\bibitem[Zhou et~al\mbox{.}(2026)]%
        {zhou2026finefake}
\bibfield{author}{\bibinfo{person}{Ziyi Zhou}, \bibinfo{person}{Xiaoming
  Zhang}, \bibinfo{person}{Litian Zhang}, \bibinfo{person}{Jiacheng Liu},
  \bibinfo{person}{Erik Cambria}, {and} \bibinfo{person}{Chaozhuo Li}.}
  \bibinfo{year}{2026}\natexlab{}.
\newblock \showarticletitle{{FineFake}: A Knowledge-Enriched Dataset for
  Fine-Grained Multi-Domain Fake News Detection}.
\newblock \bibinfo{journal}{\emph{Information Fusion}}  \bibinfo{volume}{132}
  (\bibinfo{year}{2026}), \bibinfo{pages}{104253}.
\newblock
\href{https://doi.org/10.1016/j.inffus.2026.104253}{doi:\nolinkurl{10.1016/j.inffus.2026.104253}}


\bibitem[Zhu(2025)]%
        {zhu2025adaptivevitbert}
\bibfield{author}{\bibinfo{person}{HaoChen Zhu}.}
  \bibinfo{year}{2025}\natexlab{}.
\newblock \showarticletitle{AdaptiveViTBERT: A Multimodal Fake News Detection
  Model Integrating Dynamic Gating and Missing Modality Compensation}. In
  \bibinfo{booktitle}{\emph{2025 6th International Conference on Machine
  Learning and Computer Application (ICMLCA)}}. IEEE,
  \bibinfo{pages}{1082--1088}.
\newblock


\end{thebibliography}

\appendix

\end{document}